\documentclass[
10pt, twocolumn, letterpaper, 
]{article}
\usepackage[final]{cvpr} 
\usepackage[accsupp]{axessibility}
\usepackage[dvipsnames]{xcolor}
\usepackage{algorithm, algorithmic}
\usepackage{setspace}
\usepackage{multirow}
\usepackage{enumitem}
\usepackage{amsmath}
\definecolor{cvprblue}{rgb}{0.21, 0.49, 0.74}
\usepackage[pagebackref, breaklinks, colorlinks, citecolor=cvprblue]{hyperref}
\def\paperID{14579} 
\def\confName{CVPR}
\def\confYear{2024}

\title{Efficiently Assemble Normalization Layers and Regularization for\\Federated Domain Generalization}

\author{Khiem Le$^1$, Long Ho$^2$, Cuong Do$^2$, Danh Le-Phuoc$^3$, Kok-Seng Wong$^{2,}\thanks{ Corresponding author: \url{wong.ks@vinuni.edu.vn}}$\\
$^1$ Department of Computer Science and Engineering, University of Notre Dame, IN, USA\\
$^2$ College of Engineering and Computer Science, VinUniversity, Hanoi, Vietnam\\
$^3$ Open Distributed Systems, Technical University Berlin, Berlin, Germany\\
{\tt\small kle3@nd.edu, }{\tt\small danh.lephuoc@tu-berlin.de, }{\tt\small \{long.ht, cuong.dd, wong.ks\}@vinuni.edu.vn}
}

\begin{document}
\maketitle

\begin{abstract}
Domain shift is a formidable issue in Machine Learning that causes a model to suffer from performance degradation when tested on unseen domains. Federated Domain Generalization (FedDG) attempts to train a global model using collaborative clients in a privacy-preserving manner that can generalize well to unseen clients possibly with domain shift. However, most existing FedDG methods either cause additional privacy risks of data leakage or induce significant costs in client communication and computation, which are major concerns in the Federated Learning paradigm. To circumvent these challenges, here we introduce a novel architectural method for FedDG, namely gPerXAN \footnote{\href{https://github.com/lhkhiem28/gPerXAN}{https://github.com/lhkhiem28/gPerXAN}}, which relies on a normalization scheme working with a guiding regularizer. In particular, we carefully design \textbf{Per}sonalized e\textbf{X}plicitly \textbf{A}ssembled \textbf{N}ormalization to enforce client models selectively filtering domain-specific features that are biased towards local data while retaining discrimination of those features. Then, we incorporate a simple yet effective regularizer to \textbf{g}uide these models in directly capturing domain-invariant representations that the global model’s classifier can leverage. Extensive experimental results on two benchmark datasets, i.e., PACS and Office-Home, and a real-world medical dataset, Camelyon17, indicate that our proposed method outperforms other existing methods in addressing this particular problem. 
\end{abstract}

\section{Introduction}
Over the past few decades, Machine Learning (ML) has demonstrated remarkable achievements across diverse areas such as Computer Vision, Natural Language and Speech Processing, or Robotics \cite{Applications}. In general, most ML models rely on an over-simplified assumption, i.e., the training and testing data are independent and identically distributed, which does not always reflect real-world practices. In practical scenarios where the distribution of testing data diverges from that of training data, the performance of ML models often drops catastrophically due to the domain shift issue \cite{Dataset-Shift}. Additionally, obtaining or identifying the testing data before model deployment can be challenging in numerous applications. For instance, in biomedical applications where data characteristics vary across different equipment and institutions, gathering data from all potential domains in advance is impractical. Therefore, it is essential to have a solution that can improve the generalization capability of such ML models to adapt effectively to unseen domains. 

Domain Generalization (DG) has been proposed to address the challenge of training ML models using data from single or multiple source domains with the expectation that these models will perform well on unseen domains \cite{Generalization-Survey}. The majority of existing DG methods fall under the category of \textbf{\textit{domain-invariant representation learning}} approach \cite{alignment-1, alignment-2, alignment-3, alignment-4, alignment-5}. This approach relies on a broadly acknowledged assumption that each domain contains its own domain-specific features, which are biased towards spurious relations in the data, and that all domains share domain-invariant features, which are general and robust to any unseen domains. From this assumption, previous works propose methods that remove domain-specific features and distill domain-invariant features to achieve the generalization ability. Alternative approaches for DG encompass \textit{data augmentation} \cite{augmentation-1, augmentation-2, augmentation-3, augmentation-4}, which involves exposing models to artificially generated domains, and \textit{meta-learning} \cite{meta-1, meta-2, meta-3}, an approach that emulates the domain shift during the meta-training phase. However, most methods require a centralized setting where all source domains are collected together. Consequently, these methods cannot be readily expanded to decentralized settings. 

\begin{figure*}[!t]
    \centering
    \includegraphics[width=1.0\linewidth]{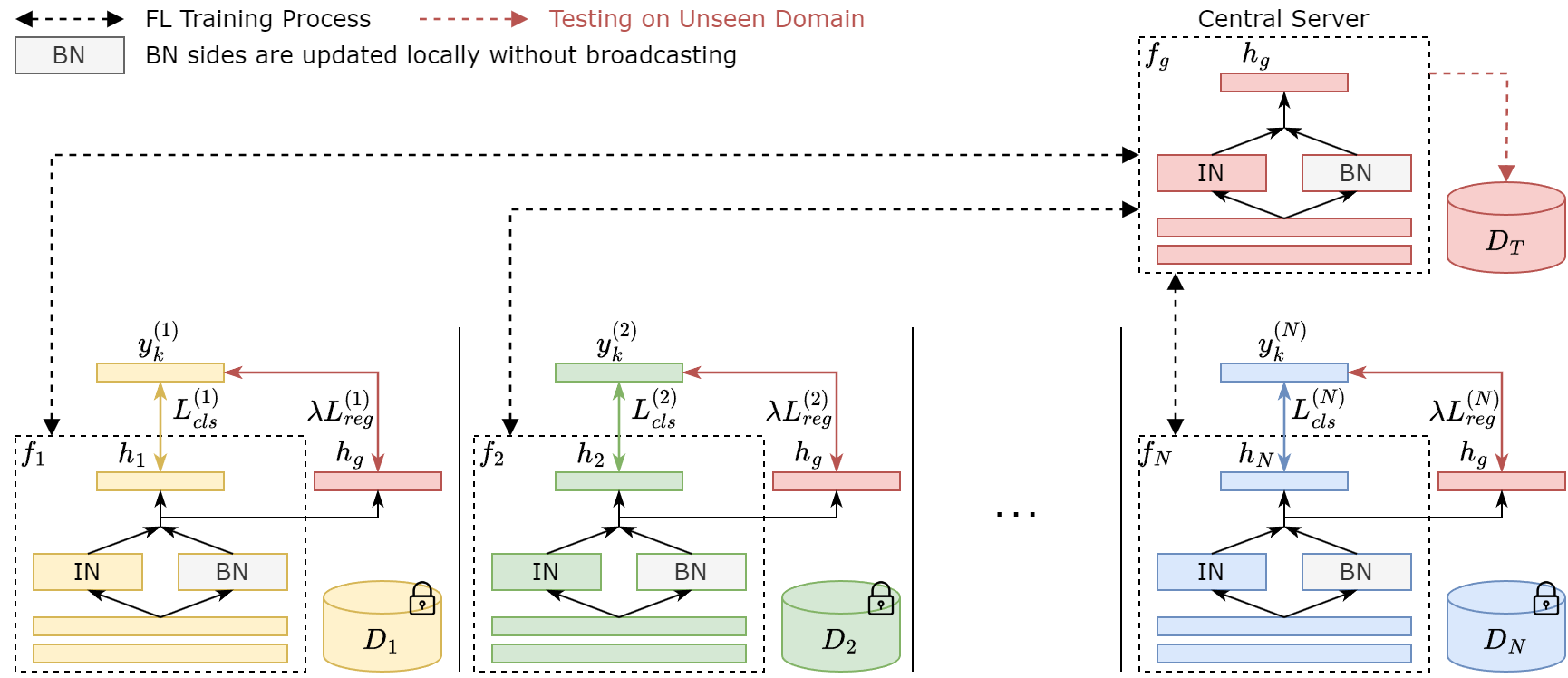}
    \caption{An overview of our proposed gPerXAN method for solving the FedDG problem. }
    \label{fig-gPerXAN}
\end{figure*}

Federated Learning (FL) \cite{FedAvg} is an emerging decentralized learning paradigm widely adopted in various applications to cope with the increasing privacy concerns of data centralization \cite{GDPR}. Specifically, the paradigm works in a way that each client learns from their data and only aggregates local models’ parameters at a certain frequency at the central server to generate a global model. Notably, all data samples are kept within each client during the FL training process. Due to the nature of data decentralization, where each client owns a single source domain, as illustrated in Figure \ref{fig-gPerXAN}, the FL paradigm poses further significant challenges for DG and limits the applicability of available DG methods. There have been some early attempts to address the DG problem in the FL scenario. For instance, \citet{ELCFS} introduces a method that allows clients to share their image data in the frequency space with each other, thus relatively recovering the centralization process at each client. Similarly, \citet{CCST} introduces another method that extracts and exchanges the style of local images among all clients. It is evident that these initial efforts employ a strategy that necessitates the sharing of partial client data, thereby compromising the data privacy constraints of FL to a certain extent. Although they show promising results, these methods can be overly complicated to implement in practice and lead to additional privacy risks during the FL training process. 

To address the aforementioned challenges, this paper introduces a novel architectural method for domain-invariant representation learning within the FL framework. The proposed method enhances the generalization ability while upholding the fundamental privacy principles of FL. Based on the effectiveness of discarding domain-specific information from learned features \cite{alignment-3, alignment-6}, we properly assemble Instance Normalization layers (IN) into Batch Normalization layers (BN) in well-known Convolutional Neural Networks (CNNs) using an explicit differential mixture as in Eqn (\ref{eqn-XAN}). Moreover, thanks to the explicit property, the benefit of personalization in FL \cite{Personalization-1, Personalization-2} can be incorporated into the normalization scheme using local BN sides. Specifically, during the FL training process, while IN sides are globally aggregated along with other model parameters, BN sides are updated locally without broadcasting. In addition, we argue that only relying on the ability to filter domain-specific features of IN while lacking guidance to distill domain-invariant representations directly might lead to suboptimal performance. Based on this observation, we introduce a simple yet highly effective regularization term to \textit{guide} client models to directly capture domain-invariant representations that can be used by the global model’s classifier, which is aggregated from client models’ classifiers. 

To summarize, our main contributions in this paper are highlighted as follows: 
\begin{itemize}[leftmargin=1.2em, topsep=0.08em, noitemsep]
    \item Different from existing methods for DG in the FL scenario, we propose a novel method that concentrates on a personalized normalization scheme, global IN while local BN, for filtering domain-specific features and fully respecting the privacy-preserving principles of FL. 
    \item Furthermore, we propose a simple yet effective regularization term to introduce clear guidance to client models for directly capturing domain-invariant representations, further improving performance on unseen domains. 
    \item Finally, we conduct extensive experiments on two benchmark datasets, i.e., PACS and Office-Home, and a real-world medical dataset, Camelyon17, where our proposed method outperforms existing relevant ones. 
\end{itemize}

\section{Related Work}
\label{sec-literature}

\subsection{Domain Generalization}
Domain Generalization is a challenging task requiring models to perform well on unseen domains. One dominant approach is \textbf{\textit{domain-invariant representation learning}}, which aims to minimize the discrepancy among source domains, assuming that the resulting representation will be domain-invariant and generalize well on testing domains. Along this track, \citet{muandet2013domain} proposes to reduce the domain dissimilarity by using Maximum Mean Discrepancy. \citet{alignment-3} combines IN and BN in popular CNNs to reap the benefits of removing domain-specific features while maintaining the ability to capture discriminative features. Meanwhile, \citet{alignment-6} discloses that combining these normalization layers in a switchable mechanism can yield better performance. Another approach to DG is \textit{data augmentation}, which augments source domains to a broader span of the training data space, enlarging the possibility of covering the span of the data in the testing domain. \citet{augmentation-3} mixes styles of training instances, resulting in novel domains being synthesized implicitly. Recently, the \textit{meta-learning} approach has drawn increasing attention from the DG community. \citet{meta-1} proposes MetaReg that learns a regularization function, particularly for the network classification layer, while excluding the feature extractor. However, the majority of these methods rely on having access to a diverse set of source domains, making them not applicable to decentralized settings. 

\subsection{Federated Domain Generalization}
While there is a growing interest in addressing the DG problem within the FL framework, particularly in scenarios where each client possesses a single source domain, the existing literature on this subject remains relatively sparse. This FedDG problem was initially introduced by \citet{ELCFS}, who then proposed ELCFS that involves exchanging amplitude information in the frequency space among clients and leveraging episodic learning to further enhance performance. \citet{CCST} proposes a similar mechanism CCST that extracts and exchanges the overall domain style of local images among all clients. Unfortunately, these early works require sharing partial information about the local data, which can be viewed as a form of data leakage and is undesirable in the FL setting. Moreover, these methods include broadcasting partial information and interpolating this information into new training data, which adds significant extra cost to communication and computation during the FL training process. Besides, \citet{COPA} proposes an architectural method COPA in which the global model consists of a domain-invariant representation extractor and an ensemble of domain-specific classifiers. \citet{FedKA} introduces FedKA that employs a server-side voting mechanism that generates target domain pseudo-labels based on the consensus from clients to facilitate global model fine-tuning. Recently, \citet{GA} proposes a novel global objective incorporating a variance reduction regularizer to encourage fairness, and then FedDG-GA is proposed to optimize this objective by dynamically calibrating the aggregation weights. Although relaxing from the data leakage issue, these methods also cause a large additional consumption of resources when the number of source domains and output classes increases. In this paper, we introduce an alternative architectural method that shares only the model update information during training. This ensures maximal data privacy and circumvents the communication and computation overhead issue mentioned earlier while achieving competitive results. In Table \ref{tab-comparison}, we highlight the advantages of our method in comparison to previous ones. 

\begin{table}[!t]
\centering
\small
\setlength{\tabcolsep}{3.65pt}
\renewcommand{\arraystretch}{0.98}
\begin{tabular}{@{}lccc@{}}
\toprule
\multirow[t]{2}{*}{Method} & \multirow[t]{2}{*}{Privacy Risk} & \multicolumn{2}{c}{Additional Cost} \\ \cmidrule(l){3-4} 
                        &                       & Communication         & Computation           \\ \midrule
ELCFS                   & \checkmark            & \checkmark            & \checkmark            \\
CCST                    & \checkmark            & \checkmark            & \checkmark            \\
COPA                    & $\boldsymbol{\times}$ & \checkmark            & \checkmark            \\
FedDG-GA                & $\boldsymbol{\times}$ & $\boldsymbol{\times}$ & \checkmark            \\ \midrule
\textbf{gPerXAN (Ours)} & $\boldsymbol{\times}$ & $\boldsymbol{\times}$ & $\boldsymbol{\times}$ \\ \bottomrule
\end{tabular}
\caption{An advantage comparison of different methods. }
\label{tab-comparison}
\end{table}

One related research field to FedDG is Federated Domain Adaptation (FedDA). While both fields aim to maximize the model performance on unseen domains using existing source domains, FedDA can access the target domains while FedDG cannot see those data during training. Leveraging this assumption, FedDA methods such as FADA \cite{peng2019federated} or FMTDA \cite{yao2022federated}, are typically able to align target and source features. However, this specific assumption also makes FedDA methods are inapplicable to FedDG. 

\subsection{Personalized Federated Learning}
Another related and orthogonal line of work is Personalized Federated Learning (pFL). pFL aims to learn personalized models for different clients to tackle distribution shifts across client data, known as data heterogeneity. \citet{li2020federated} utilizes a regularization term in the local loss function so that the clients’ trained models will not significantly differ from the global model. On the other side, \citet{Personalization-1} and \citet{Personalization-2} mitigate the heterogeneous distributions by only loading a subset of the global model’s parameters rather than loading the entire model at each training round. Although related in terms of overcoming distribution shifts across clients, pFL focuses on improving the performance of participating clients, whereas our considered field FedDG focuses on improving the performance of unseen clients with unseen domains. 

\section{Methodology}

\noindent
\textbf{Problem Formulation}. First, we denote $X$ and $Y$ as the input space and the label space, respectively, of a specific task $\mathcal{T}$. In the standard FL setting, $N$ clients $\{c_i\}_{i=1}^N$ are involved in collaboratively constructing a global model $f_g$ for solving the task, where each client $c_i$ owns a dataset $D_i = \{(x_k, y_k)\}_{k=1, |D_i|}$ which is associated to a specific domain defined by a joint distribution $P_{X, Y}^{(i)}$. Importantly, scattered datasets $\{D_i\}_{i=1}^N$ across clients satisfy: 
\begin{itemize}[leftmargin=1.0em, topsep=0.5em]
   \item $P_{X, Y}^{(i)} \neq P_{X, Y}^{(j)}$ with $1 \le i, j \le N$ and $i \neq j$
   \item $P_{Y| X}^{(i)}    = P_{Y| X}^{(j)}$ with $1 \le i, j \le N$ and $i \neq j$
\end{itemize}
The objective of FedDG is leveraging scattered datasets to construct a model $f_g$ that can directly generalize to the unseen dataset $D_U$ with an unseen domain, which means: 
\begin{itemize}[leftmargin=1.0em, topsep=0.5em]
   \item $P_{X, Y}^{(U)} \neq P_{X, Y}^{(i)}$ with $1 \le i \le N$
   \item $P_{Y| X}^{(U)}    = P_{Y| X}^{(i)}$ with $1 \le i \le N$
\end{itemize}
To this end, $N$ clients communicate with a central server for $T$ rounds. At each round, every client $c_i$ receives the same global model $f_g$ from the server and updates $f_g$ with their local dataset $D_i$ for $E$ epochs to establish its local model $f_i$. The server then collects all trained models and aggregates them to update the global model. This process repeats until the global model converges. In this work, we consider the most popular FL framework, FedAvg \cite{FedAvg}, which aggregates client models as:
\begin{equation}
\label{eqn-FedAvg}
f_g = \sum_{i=1}^{N} \frac{|D_i|}{\sum_{1\le j\le N}|D_j|} f_i
\end{equation}

\noindent
\textbf{Challenges}. In the general spirit of DG, a model is expected to extensively explore multiple source domains to achieve domain-invariance in its learned latent representation. However, under the FL setting, each client is restricted to accessing only its own local data, which constrains to make full use of source domains and, consequently, limits learning of generalizable representation. Moreover, sharing data or even partial information about data among clients poses additional privacy risks and introduces communication and computation costs during the FL training process. 

\subsection{Normalization Scheme}

\noindent
\textbf{eXplicitly Assembled Normalization}. The first step towards solving the highlighted challenges, we introduce a novel normalization scheme called  eXplicitly Assembled Normalization (XAN), which combines IN and BN as: 
\begin{equation}
\label{eqn-XAN}
\hat{h} = w_{in}(\gamma_{in}\frac{h - \mu_{in}}{\sqrt{\sigma_{in}^2 + \epsilon}} + \beta_{in})
        + w_{bn}(\gamma_{bn}\frac{h - \mu_{bn}}{\sqrt{\sigma_{bn}^2 + \epsilon}} + \beta_{bn}), 
\end{equation}
where $h$, $\hat{h} \in \mathrm{R}^{B\text{x}C\text{x}W\text{x}H}$ are layer input and output activations, which are 4D tensors with dimensions of batch size $B$, number of channels $C$, width $W$ and height $H$. Meanwhile, $\mu$ and $\sigma^2$ are means and variances captured by the normalization layers, respectively, $\gamma$ and $\beta$ are affine parameters, and $\epsilon$ is for numerical stability. 

In essence, XAN is a sophisticated mixture mechanism of IN and BN to replace BN in the feature extractors of CNNs. In particular, $w_{in}$ and $w_{bn}$ are ratios to weight the mixture that allow the model to switch between IN and BN. These parameters are randomly initialized and optimized along with other model parameters during training in an end-to-end manner. Unlike previous works \cite{alignment-7, alignment-6}, which uses an implicit mechanism to combine computed statistics of IN and BN, i.e., means and variances, XAN uses an explicit mechanism that combines the output activations of two normalization layers, this provides the model with a unique ability to separate IN from BN completely. Note that IN has shown great success in style transfer tasks \cite{AdaIN} as it allows discarding the variability of visual styles such as object colors or textures from the content of images. This property makes IN beneficial for solving the DG problem. However, directly using IN to replace the conventional BN leads to the loss of discrimination in the learned features, resulting in performance degradation in classification tasks. 

\noindent
\textbf{Personalized eXplicitly Assembled Normalization}. To accomplish our proposed normalization scheme, Personalized eXplicitly Assembled Normalization (PerXAN), we make a subtle modification to the conventional FedAvg \cite{FedAvg} framework based on the explicit property of the above XAN. Specifically, during the FL training process, while IN sides of XAN layers are globally aggregated along with other model parameters, BN sides are updated locally, which means that parameters of BN sides are excluded from the broadcasting steps from the server. Notably, in inference time, the global model is generated by averaging all model parameters of clients. Our modification is motivated by a common observation that FedAvg normally results in poor convergence and performance in the presence of domain heterogeneity across clients \cite{FedAvg-Convergence, zhuangfedwon} with the major reason being that client models forget the acquired knowledge from previous rounds after aggregated \cite{FedAvg-Forgetting}. The leading solution is to personalize a subset of the model, which benefits clients in learning better from their local data \cite{Personalization-1, Personalization-2}. Moreover, based on another finding that BN plays a crucial role in dealing with the domain shift issue in the centralized paradigm \cite{BN-1, BN-2}, BN sides in XAN layers are finally made to be personalized. PerXAN is illustrated in Figure \ref{fig-gPerXAN}. 

\subsection{Regularization as Guidance}
With the proposed normalization scheme PerXAN in place, now, we are ready to introduce our \emph{"Regularization as Guidance"}. It targets to induce clear guidance to client models through a regularizer, which brings an effect of domain alignment \cite{sun2017correlation, alignment-2}, and guides these models to capture domain-invariant representations directly. We hypothesize that this will address a drawback of current DG methods that are later verified by our experiments in Section \ref{experiments}. The drawback is from DG methods that take advantage of the IN’s function \cite{alignment-3, alignment-6} but do not actually equip the model with the capability of capturing domain-invariant features even if they demonstrated promising performance. Instead, it is only hoped that domain-invariant features would be distilled through achieving the goal of removing domain-specific features. This indirect learning purpose might affect the model’s learning efficacy, especially in the FL setting, where each client only owns a separate single source domain. Specifically, we assume a classification model $f$, either the global model $f_g$ or a client model $f_i$, comprises a feature extractor $g$ and a classifier head $h$. At each client $c_i$, during local training, the client model $f_i$ is optimized on the local dataset $D_i$ using the following loss function: 
\begin{align}
L_i &= L_{cls}^{(i)}(f_i; D_i) + \lambda L_{reg}^{(i)}(g_i, h_g; D_i) \label{eqn-Reg-1} \\
    &= \sum_{k=1}^{|D_i|} \ell(f_i(x_k^{(i)}), y_k^{(i)}) + \lambda \ell(h_g(g_i(x_k^{(i)})), y_k^{(i)}), \label{eqn-Reg-2}
\end{align}
where $\ell$ is the base loss function, which is usually cross-entropy in classification tasks, and $\lambda$ is a hyper-parameter to control the significance of the regularizer. In a deeper look, by freezing $h_g$ during local training, the auxiliary loss term $L_{reg}$ forces client models to arrive at representations that can be made use for classification by the same global classifier, hence, producing an alignment effect on these representations \cite{sun2017correlation, alignment-2}. Moreover, our proposed regularizer can also be interpreted as an implicit form of matching global knowledge to clients’ knowledge, which has been demonstrated to bring performance gain \cite{ELCFS, CCST} for FedDG. 

Due to its orthogonality, our proposed regularizer can be easily integrated into various methods, as shown in Algorithm \ref{alg-gPerXAN}. Specifically, our experiments in Section \ref{experiments} verify its compatibility with the normalization scheme that justifies our motivation. Moreover, by using only the global model’s classifier to regularize feature extractors, gPerXAN saves major communication and computation resources, as well as memory usage at clients compared to others, which use an ensemble of classifiers \cite{COPA} or entire global model \cite{GA}. 

\begin{algorithm}[!t]
\setstretch{1.18}
\caption{gPerXAN}
\label{alg-gPerXAN}
\begin{algorithmic}[1]
\STATE \textbf{Input}: A model $f$ uses PerXAN to replace BN layers in the feature extractor. $N$ clients with their local datasets $\{D_i\}_{i=1}^N$. Notably, $f^{(t)}$ and $f^{(l; t)}$ is the model $f$ and its $l^{th}$ layer at the communication round $t$, respectively. 
\STATE \textbf{Initialization}: $f_g^{(0)} \gets f$
\FOR{each round $t = 1, 2, 3, \dots, T$}
\FOR{each client $i = 1, 2, 3, \dots, N$ \textbf{in parallel}}
\FOR{each layer in $f_i^{(t)}$}
\IF{$f_i^{(l; t)}$ is not BN}
\STATE $f_i^{(l; t)} \gets f_g^{(l; t)}$
\ENDIF
\ENDFOR
\STATE $f_i^{(t)} \gets \textbf{LocalTraining}(f_i^{(t)}, h_g; D_i)$
\ENDFOR
\STATE $f_g^{(t)} = \sum_{i=1}^{N} \frac{|D_i|}{\sum_{1\le j\le N}|D_j|} f_i^{(t)}$ \hspace*{\fill} // Eqn (\ref{eqn-XAN})
\ENDFOR
\STATE \textbf{return}: $f_g^{(T)}$
\STATE $\textbf{LocalTraining}(f_i,h_g; D_i)$:
\FOR{each epoch $e = 1, 2, 3, \dots, E$}
\STATE $L_i = L_{cls}^{(i)}(f_i; D_i) + \lambda L_{reg}^{(i)}(g_i, h_g; D_i)$ \hspace*{\fill} // Eqn (\ref{eqn-Reg-1})
\STATE $f_i \gets f_i -\eta \nabla L_i$
\ENDFOR
\STATE \textbf{return}: $f_i$
\end{algorithmic}
\end{algorithm}

\section{Experiments and Results}
\label{experiments}

\subsection{Datasets}
To evaluate the proposed method, we perform experiments on two standard DG benchmark datasets, i.e., PACS \cite{PACS} and Office-Home \cite{Office-Home}, and a real-world medical image dataset, Camelyon17 \cite{Camelyon17}, consisting of various sub-datasets that are considered as domains. Specifically, PACS is composed of 4 domains with large discrepancies from diverse image colors and textures, \textit{Photo} (P), \textit{Art painting} (A), \textit{Cartoon} (C), and \textit{Sketch} (S). Each domain contains 7 categories, with 9,991 images in total. Office-Home is also composed of 4 domains but with smaller discrepancies from various backgrounds and camera viewpoints, \textit{Product} (P), \textit{Art} (A), \textit{Clipart} (C), and \textit{Real-world} (R). Each domain contains a more extensive label set of 65 categories with 15,588 images. Camelyon17 is a binary tumor classification dataset containing 455,964 histology images with stains from 5 \textit{different hospitals} worldwide. These datasets are at different difficulty levels and are commonly used in the literature. 

\begin{table*}[!t]
\centering
\setlength{\tabcolsep}{7.55pt}
\renewcommand{\arraystretch}{1.18}
\begin{tabular}{@{}lrrrrrrrrrrr@{}}
\toprule
\multirow[t]{2}{*}{Method} & \multicolumn{5}{r}{PACS}                                                                                                        & \multicolumn{5}{r}{Office-Home}                                                                                                 &                      \\ \cmidrule(lr){2-11}
                        & P                    & A                    & C                    & S                    & \multicolumn{1}{r|}{Avg}            & P                    & A                    & C                    & R                    & \multicolumn{1}{r|}{Avg}            & Avg                  \\ \midrule
FedAvg (Baseline)       & 95.21                & 82.23                & 78.20                & 73.56                & \multicolumn{1}{r|}{82.30}          & 76.53                & 65.97                & 55.40                & 78.01                & \multicolumn{1}{r|}{68.98}          & 75.64                \\ \midrule
FedAvg w/ MixStyle      & 95.93                & 85.99                & 80.03                & 75.46                & \multicolumn{1}{r|}{84.35}          & 75.87                & 62.09                & 57.92       & 77.48                & \multicolumn{1}{r|}{68.34}          & 76.35                \\
FedAvg w/ RSC           & 95.21                & 83.15                & 78.24                & 74.62                & \multicolumn{1}{r|}{82.81}          & 75.26                & 62.34                & 50.79                & 77.46                & \multicolumn{1}{r|}{66.46}          & 74.63                \\ \midrule
ELCFS                   & 96.23                & 83.94                & 79.27                & 73.30                & \multicolumn{1}{r|}{83.19}          & 76.83                & 66.32                & 55.63                & 78.12                & \multicolumn{1}{r|}{69.23}          & 76.21                \\
CCST                    & 96.65                & 88.33       & 78.20                & 82.90                & \multicolumn{1}{r|}{86.52}          & 76.61                & 66.35                & 52.39                & 80.01                & \multicolumn{1}{r|}{68.84}          & 77.68                \\
COPA                    & 95.62                & 84.80                & 80.28                & 82.86                & \multicolumn{1}{r|}{85.89}          & 75.82                & 62.27                & 56.04                & 78.72                & \multicolumn{1}{r|}{68.21}          & 77.05                \\
FedDG-GA                & 96.80                & \textbf{86.91}                & 81.23                & 82.74                & \multicolumn{1}{r|}{86.92}          & 77.23                & 65.10                & \textbf{58.29}                & 78.80                & \multicolumn{1}{r|}{69.86}          & 78.42                \\ \midrule
\textbf{gPerXAN (Ours)} & \textbf{97.27}       & 86.52                & \textbf{84.68}       & \textbf{83.28}       & \multicolumn{1}{r|}{\textbf{87.94}} & \textbf{78.91}       & \textbf{67.24}       & 57.75                & \textbf{80.15}       & \multicolumn{1}{r|}{\textbf{71.01}} & \textbf{79.48}       \\ \bottomrule
\end{tabular}
\caption{Accuracy comparison on the PACS and Office-Home datasets in the leave-one-domain-out setting. }
\label{tab-main-results}
\end{table*}

\subsection{Experimental Settings}
\noindent
\textbf{Evaluation}. For a fair comparison purpose, we follow the common leave-one-domain-out evaluation protocol as considered in \cite{ELCFS, CCST, GA}. In particular, we sequentially choose one domain as the unseen domain, train the model on all remaining domains where a single domain is treated as a client, and evaluate the trained model on the chosen domain. For PACS and Office-Home datasets, we split 90\% of the data of each source client as the training set and 10\% of that as the validation set, while for unseen clients, the entire data is used for testing. For the large-scale Camelyon17 dataset, the ratios of training set and validation set at each source client are 80\% and 20\%, respectively, while the entire data is used for testing at unseen clients. In all experiments, we report the test accuracy on each unseen client by using the best validation model selected based on the average of accuracies on validation sets of source clients. The reported numerical values are averaged over 3 runs. Finally, we straightforwardly compare the proposed method with the vanilla FedAvg \cite{FedAvg}. Two centralized DG methods, which are free from the requirement of data centralization, MixStyle \cite{augmentation-3} and RSC \cite{alignment-8}, are also evaluated under the integration into the FedAvg framework. Furthermore, we directly compare our method with state-of-the-art relevant ones that address the problem of DG in the FL setting as discussed in Section \ref{sec-literature}, including ELCFS \cite{ELCFS}, CCST \cite{CCST}, COPA \cite{COPA}, and FedDG-GA \cite{GA}. 

\begin{table}[!t]
\centering
\setlength{\tabcolsep}{4.07pt}
\renewcommand{\arraystretch}{1.18}
\begin{tabular}{@{}lrrrrrr@{}}
\toprule
\multirow[t]{2}{*}{Method} & \multicolumn{6}{r}{Camelyon17}                                                                \\ \cmidrule(l){2-7} 
                        & H1            & H2            & H3            & H4            & H5            & Avg           \\ \midrule
FedAvg (Baseline)       & 97.0          & 91.8          & 89.9          & 94.2          & 81.0          & 90.8          \\ \midrule
FedAvg w/ MixStyle      & 91.1          & 85.5          & 86.2          & 93.3          & 87.9          & 88.8          \\
FedAvg w/ RSC           & 90.6          & 90.6          & 88.3          & 94.5          & \textbf{93.3}          & 91.5          \\ \midrule
ELCFS                   & 92.9          & 90.6          & 89.9          & 93.2          & 89.9          & 91.3          \\
CCST                    & 91.5          & 90.2          & 87.3          & 94.6          & 91.6          & 91.0          \\
COPA                    & 93.2          & 90.9          & 92.2          & 93.6          & 90.2          & 92.0          \\
FedDG-GA                & \textbf{97.2}          & 90.7          & 91.0          & 92.3          & 90.5          & 92.3          \\ \midrule
\textbf{gPerXAN (Ours)} & 96.5          & \textbf{92.2}          & \textbf{95.1}          & \textbf{94.7}         & 91.9          & \textbf{94.1}          \\ \bottomrule
\end{tabular}
\caption{Accuracy comparison on the Camelyon17 dataset. }
\label{tab-medical-results}
\end{table}

\noindent
\textbf{Implementation Details}. We present architectural details and hyper-parameter values used for experiments in the paper. Following \cite{CCST, GA}, for PACS and Office-Home datasets, we use the image size of 224x224 pixels and ResNet-50 \cite{ResNet} pre-trained on ImageNet as the backbone for the feature extractor and a linear layer as the classifier. BN layers in the first four and first two blocks in feature extractors are replaced with our PerXAN, respectively. For the Camelyon17 dataset, input images are resized to 96x96 pixels, and DenseNet-121 \cite{DenseNet} is used instead of ResNet-50, BN layers in the first dense block and the following transition block are replaced by PerXAN. Notably, parameters related to BN layers are initialized with ImageNet pre-trained weights. Client models in all experiments are optimized by an SGD optimizer with a learning rate of 2e-3 for 100 communication rounds (i.e. $T$ = 100) with one local update epoch at clients (i.e. $E$ = 1). During the training, simple data augmentation techniques are applied including random horizontal flipping and color jittering. The hyper-parameter of our regularization term $\lambda$ is searched in the range [0, 1] with a step of 0.25. The MixStyle \cite{augmentation-3} and RSC \cite{alignment-8} can be directly integrated into the FedAvg without further modifications. All settings of other compared methods, i.e., ELCFS \cite{ELCFS}, CCST \cite{CCST}, COPA \cite{COPA}, and FedDG-GA \cite{GA} are chosen based on corresponding papers. 

\subsection{Main Results}
Table \ref{tab-main-results} presents the quantitative results for different testing domains on the PACS and Office-Home datasets. Particularly, each result column shows the test accuracy of the global model on the domain of the column name. First of all, we can empirically verify that centralized DG methods such as MixStyle \cite{augmentation-3} and RSC \cite{alignment-8} are inconsistent under the FedAvg framework and even harmful in certain cases compared to the baseline. Since these methods are designed for the centralized setting, they might need access to inter-domain knowledge to perform well, which is unavailable in the FL setting. In comparing state-of-the-art FedDG methods designed for the FL scenario, methods that share partial client data information, i.e., ELCFS \cite{ELCFS} and CCST \cite{CCST}, show impressive results on the PACS dataset. However, they do not significantly affect the Office-Home dataset. A similar pattern is found with the architectural method COPA \cite{COPA} and FedDG-GA \cite{GA}. Meanwhile, on the PACS and Office-Home datasets, our proposed method, gPerXAN achieves average accuracies across unseen clients of 87.94\% and 71.01\%, which are 1.02\% and 1.15\% better than the second-best ones, respectively. Although most of the considered methods can perform better than the baseline FedAvg in most cases, our method demonstrates a significant boost over others on both two standard benchmarks. 

In addition to standard PACS and Office-Home datasets, we further evaluate our proposed method on a medical imaging dataset Camelyon17 as presented in Table \ref{tab-medical-results}. On this real-world benchmark, we can observe that ELCFS \cite{ELCFS} and CCST \cite{CCST} show considerably inferior performance compared to architectural ones, i.e., COPA \cite{COPA} and gPerXAN. This might be due to sophisticated and sensitive features in medical images that are more challenging to extract and interpolate than conventional datasets \cite{ML-Med}, leading to inefficiency in these information-sharing-based methods. Notably, although it yields impressive performance, COPA \cite{COPA} involves several other advanced techniques, such as RandAugment \cite{RandAugment}, making the performance gain hard to justify. Meanwhile, our method achieves an average accuracy across unseen clients of 94.1\%, outperforming FedDG-GA \cite{GA} by approximately 2\%. In general, the above experimental results demonstrate the effectiveness of the proposed method across various applications. 

\subsection{Ablation Studies}
We conduct ablation studies on the PACS dataset to investigate the impact of each building component. Specifically, we first compare PerXAN with other centralized variants such as conventional BN, I-BN \cite{alignment-3}, and DSON \cite{alignment-6} under the same implementation details, and then our guiding regularizer is applied to ELCFS \cite{ELCFS} and CCST \cite{CCST} to verify its compatibility with the rest of the method. 

\subsubsection{Impact of the Normalization Scheme}
Table \ref{tab-Normalization} provides a quantitative comparison among considered normalization schemes where gFedAvg represents the FedAvg framework implemented with the guiding regularizer. This means that we sequentially utilize the conventional BN, I-BN \cite{alignment-3}, and DSON \cite{alignment-6} to replace the PerXAN scheme in our proposed method. From this table, we can observe that I-BN \cite{alignment-3} can yield better performance than DSON \cite{alignment-6} and the conventional BN, which are slightly comparable to each other. Meanwhile, PerXAN shows optimal performance with a significant margin compared to others and largely contributes to the whole method. 

\begin{table}[!t]
\centering
\setlength{\tabcolsep}{4.64pt}
\renewcommand{\arraystretch}{1.18}
\begin{tabular}{@{}lrrrrr@{}}
\toprule
\multirow[t]{2}{*}{Method} & \multicolumn{5}{r}{PACS}                                                           \\ \cmidrule(l){2-6} 
                        & P              & A              & C              & S              & Avg            \\ \midrule
gFedAvg w/ BN           & 95.85          & 84.39          & 78.04          & 76.42          & 83.68          \\
gFedAvg w/ I-BN         & 94.67          & 82.08          & 80.46          & 80.07          & 84.32          \\
gFedAvg w/ DSON         & 93.77          & 81.64          & 79.91          & 80.63          & 83.99          \\ \midrule
\textbf{gPerXAN (Ours)} & \textbf{97.27} & \textbf{86.52} & \textbf{84.68} & \textbf{83.28} & \textbf{87.94} \\ \bottomrule
\end{tabular}
\caption{Evaluation of different normalization schemes. }
\label{tab-Normalization}
\end{table}

\begin{figure}[!t]
    \centering
    \includegraphics[width=\linewidth]{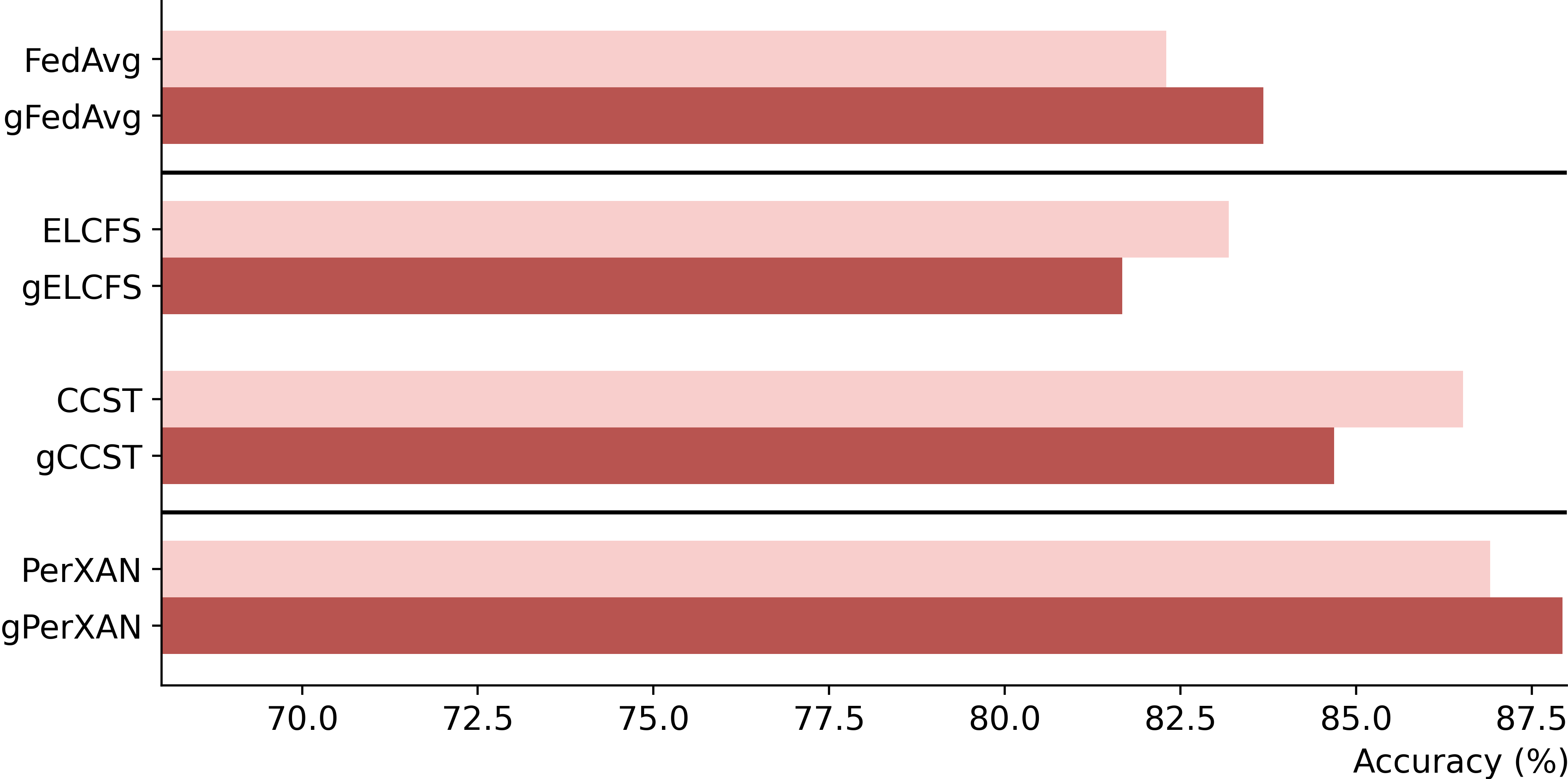}
    \caption{Contribution of the regularizer on different methods. }
    \label{fig-Regularization}
\end{figure}

\subsubsection{Impact of the Regularizer}
To better understand the proposed regularization term's contribution, we sequentially employ it in FedAvg, ELCFS \cite{ELCFS}, and CCST \cite{CCST} to observe performance changes. Figure \ref{fig-Regularization} displays averaged accuracies across unseen clients of these methods without and with the involvement of our regularizer. It is straightforward to see that while regularizer improves the performance of FedAvg and PerXAN considerably, it does not enhance ELCFS \cite{ELCFS} and CCST \cite{CCST}. In information-sharing-based methods, i.e., ELCFS \cite{ELCFS} and CCST \cite{CCST}, each client is exposed to other clients’ data information, which means that clients can access global knowledge relatively, then making the effect of matching global knowledge to clients’ knowledge redundant and harmful. Also, this ablation study empirically verifies the connection between the regularizer and normalization scheme. 

\section{Analysis and Discussion}

\begin{figure*}[!t]
    \centering
    \includegraphics[width=0.96\linewidth]{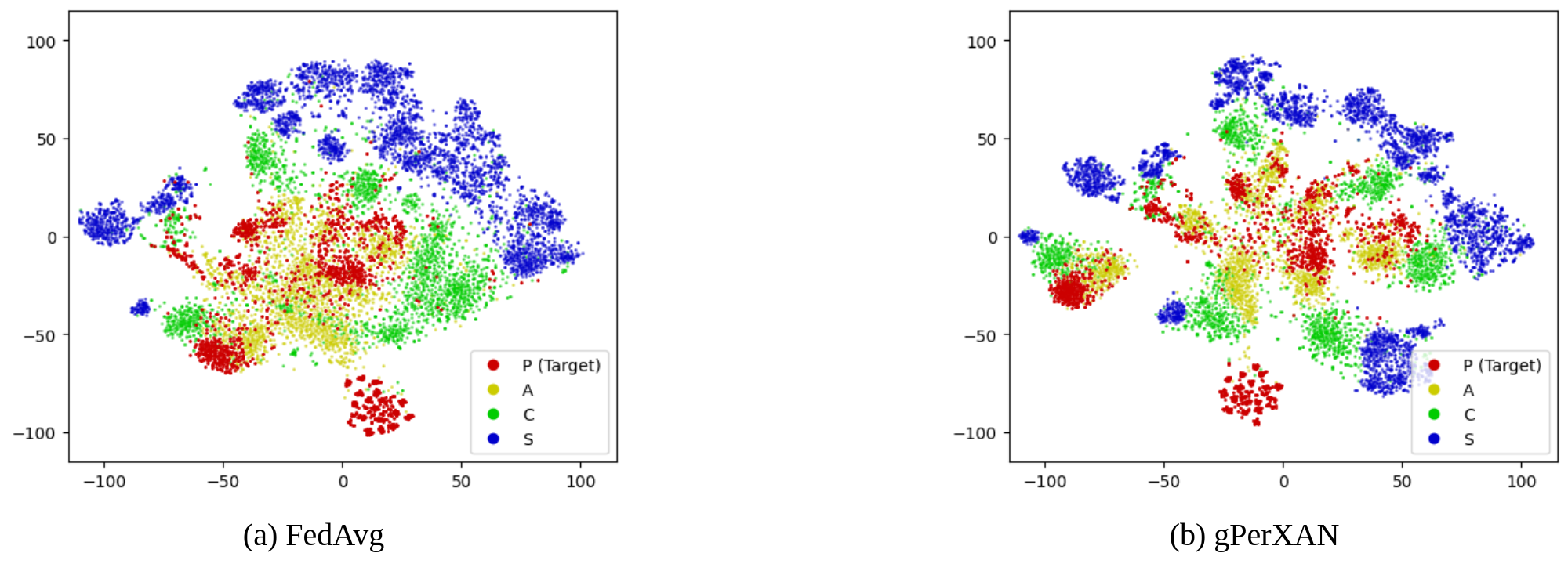}
    \caption{Visualization of representation extracted from the global model on the PACS dataset. }
    \label{fig-Analysis}
\end{figure*}

\subsection{Visualization}
We now provide an in-depth ability analysis of the proposed method via representation visualization using t-SNE \cite{t-SNE}, a common dimensionality reduction technique. As shown in Figure \ref{fig-Analysis}, we compare the representations extracted from the global model obtained from our gPerXAN and the baseline FedAvg on the PACS dataset with the testing domain, a.k.a. target domain \textit{Photo} (P). From this figure, we can observe that features derived from our method are semantically separated according to 7 categories in the PACS dataset for both source domains and the target domain. Moreover, features of each category on all domains tend to be close and grouped, demonstrating the ability of our method to distill more discriminative and domain-invariant representation, leading to a significant improvement of more than 2\% in classification accuracy on the testing domain \textit{Photo} (P).

\subsection{Privacy and Efficiency}
Data privacy is a major concern in the field of FL, which is mitigated by only sharing client models’ parameters instead of raw data. However, to resolve the FedDG problem, recently introduced methods sacrifice this principle by revealing partial information about client data. Specifically, leaking more information not only opens higher chances for attackers to perform an inversion attack \cite{inversion}, which aims to reconstruct the original data of clients, but also amplifies the risk of membership inference attack \cite{membership}, which determines if a sample was in the model's training data. Furthermore, efficiency in communication and computation is critical in FL, especially in scenarios where resource-constrained clients are involved. Despite that, the common sharing mechanism in available methods introduces unacceptable extra costs to the FL training process. For COPA \cite{COPA}, using an ensemble of domain-specific classifiers in the model architecture results in an $O(N^2)$ increase in communication and computation complexities compared to $O(N)$ as conventional, according to the number of participating clients $N$. Meanwhile, FedDG-GA \cite{GA} consumes a double of memory at clients. Compared with existing ones, in addition to bypassing the limitations described above, our proposed method is more practical in terms of implementation yet provides competitive results in extensive evaluations. 

\section{Conclusion}
In this paper, we introduce a novel architectural method, namely gPerXAN, to address the problem of FedDG. By explicitly assembling Instance Normalization layers into Batch Normalization layers in a personalized scheme and employing a simple yet effective guiding regularizer, our method allows the model to filter domain-specific features and actively distill domain-invariant representation for classification tasks. We conduct extensive experiments and in-depth analysis to quantitatively and qualitatively verify the effectiveness of our proposed method in solving this particular problem. Although our evaluation is more on cross-silo FL, our method can be easily extended to cross-device scenarios while Algorithm \ref{alg-gPerXAN} remains unchanged. Unlike existing methods, due to the independence from imaging techniques, gPerXAN can be straightforwardly extended to diverse applications. A potential limitation of our method is its certain suitability with normalization-based models. In the future, investigating other forms of regularization terms is a promising research direction due to the available room for improvements. Moreover, the vulnerability of available methods under various attacks remains underexplored. 
\section*{Acknowledgement}
This work is supported by the Deutsche Forschungsgemeinschaft, German Research Foundation under grant number 453130567 (COSMO), by the Horizon Europe Research and Innovation Actions under grant number 101092908 (SmartEdge), and by the Federal Ministry for Education and Research, Germany under grant number 01IS18037A (BIFOLD). 

\clearpage
{
\small
\bibliographystyle{cvpr}
\bibliography{refs}
}

\end{document}


\maketitle
\appendix

\section{Code Availability}
Our source code for running experiments was built originally using PyTorch \cite{PyTorch} and Flower \cite{Flower} frameworks and is attached along with supplementary material. 

\section{Additional Details for Section 4.2}
As we mentioned in the paper, the hyper-parameter of our regularization term $\lambda$ is searched in the range [0, 1] with a step of 0.25. Here we provide the results of different $\lambda$ on the PACS dataset. As shown in Table \ref{tab-app}, we sequentially set the value of $\lambda$ as 0.00, 0.25, 0.50, 0.75, 1.00 and observed that the average performance can be improved consistently with different $\lambda$. The best results are found at $\lambda$ is 0.50. Note that when $\lambda$ is set as 0.00, the guiding regularizer is not used during training, which means that gPerXAN becomes PerXAN accordingly. 

\begin{table}[!ht]
\centering
\setlength{\tabcolsep}{6.65pt}
\renewcommand{\arraystretch}{1.22}
\begin{tabular}{@{}lrrrrr@{}}
\toprule
\multirow[t]{2}{*}{$\lambda$ in gPerXAN} & \multicolumn{5}{r}{PACS}                                                           \\ \cmidrule(l){2-6} 
                        & P              & A              & C              & S              & Avg            \\ \midrule
0.00           & 96.35          & 86.28          & 83.49          & 82.01          & 86.91          \\
0.25           & 96.47          & 86.13          & 83.79          & 83.00          & 87.35          \\
0.50           & 97.27          & 86.52          & \textbf{84.68}          & 83.28          & \textbf{87.94}          \\
0.75           & \textbf{97.28}          & \textbf{86.60}          & 84.37          & \textbf{83.47}          & 87.93          \\
1.00           & 96.45          & 85.89          & 83.31          & 82.41          & 87.02          \\ \bottomrule
\end{tabular}
\caption{Impact of $\lambda$ on the PACS dataset. }
\label{tab-app}
\end{table}

\section{Additional Ablation Study}
To better understand the proposed regularization term’s contribution, we now conduct an additional comparison between our regularizer and the one from FedProx \cite{li2020federated} on enhancing the proposed normalization scheme PerXAN. As shown in Table \ref{tab-app2}, FedProx \cite{li2020federated} seems to not enhance our PerXAN. This might be due to the difference in the design purpose of regularizers. Also, we argue that our two components, the normalization scheme, and the regularizer, have a solid connection and support to each other during training, hence, gPerXAN shows better results. 

\begin{table}[!ht]
\centering
\setlength{\tabcolsep}{4.12pt}
\renewcommand{\arraystretch}{1.22}
\begin{tabular}{@{}lrrrrr@{}}
\toprule
\multirow[t]{2}{*}{Method} & \multicolumn{5}{r}{PACS}                                                           \\ \cmidrule(l){2-6} 
                        & P              & A              & C              & S              & Avg            \\ \midrule
PerXAN                  & 96.35          & 86.28          & 83.49          & 82.01          & 86.91          \\
PerXAN w/ FedProx       & 95.12          & 85.74          & 83.29          & 82.76          & 86.73          \\
\textbf{gPerXAN (Ours)} & \textbf{97.27}          & \textbf{86.52}          & \textbf{84.68}          & \textbf{83.28}          & \textbf{87.94}          \\ \bottomrule
\end{tabular}
\caption{A comparison of two regularizers on the PACS dataset. }
\label{tab-app2}
\end{table}

\section{Additional Analysis}
Regarding efficiency, we provide a more detailed analysis, which is even more solid than supporting experiments, to clarify our improvements. Here we denote $N$, $C$, and $d$ as the number of clients, number of classes, and dimension of the final feature vector in the model. Notably, we use the number of parameters as a metric for evaluating memory consumption, communication, and computation costs. Assuming training a ResNet-50 model with $R$ parameters, the table below compares the efficiency of COPA \cite{COPA}, FedDG-GA \cite{GA}, and \textbf{gPerXAN (Ours)}. The provided table clearly demonstrates our method has comparable efficiency with FedAvg \cite{FedAvg} and is much better compared to others. 
\begin{table}[!t]
\centering
\scriptsize
\setlength{\tabcolsep}{3.65pt}
\renewcommand{\arraystretch}{1.22}
\begin{tabular}{@{}llll@{}}
\toprule
\multirow[t]{2}{*}{Method} & \multirow[t]{2}{*}{Memory} & \multicolumn{2}{c}{Additional Cost} \\ \cmidrule(l){3-4} 
                        &                       & Communication         & Computation           \\ \midrule
FedAvg                    & $R$ & $R$ & $R$ \\ \midrule
COPA                    & $R+N(N \! - \! 1)Cd$ & $R+N(N \! - \! 1)Cd$ & $R+N(N \! - \! 1)Cd$ \\
FedDG-GA                & $R \times 2$ & $R$          & $R \times 2$ \\ \midrule
\textbf{gPerXAN (Ours)} & $R$          & $R$          & $R+NCd$      \\ \bottomrule
\end{tabular}
\label{tab-comparison}
\end{table}

{
\small
\bibliographystyle{cvpr}
\bibliography{refs}
}